\documentclass[10pt,conference,table]{IEEEtran}
\IEEEoverridecommandlockouts

\usepackage{soul}
\usepackage{hyperref}
\usepackage{multirow}
\usepackage{listings}
\usepackage{fancybox}
\usepackage{enumitem}
\usepackage{graphicx}
\usepackage{color}
\usepackage{array}
\usepackage{amsmath}
\usepackage{xspace}
\usepackage{url}
\usepackage{tikz}
\usepackage{caption}
\usepackage{pgfplots}
\usepackage{balance}
\usepackage{subcaption}
\pgfplotsset{compat=1.16}
\usepackage{pgf-pie}
\usetikzlibrary{pgfplots.statistics,calc}
\usepackage{algorithm}
\usepackage{algorithmic}

\usepackage{tcolorbox}
\makeatletter
\newcommand{\mybox}[1]{%
	\setbox0=\hbox{#1}%
	\setlength{\@tempdima}{\dimexpr\wd0+13pt}%
	\begin{tcolorbox}[boxrule=0.5pt, colback=white, arc=4pt,
		left=6pt,right=6pt,top=6pt,bottom=6pt,boxsep=0pt]
		#1
	\end{tcolorbox}
}

\definecolor{songcolor}{RGB}{191,191,191}

\AtBeginDocument{%
  }

\begin{document}

\title{Assessing Evaluation Metrics for Neural Test Oracle Generation}


\author{\IEEEauthorblockN{1\textsuperscript{st} Jiho Shin}
\IEEEauthorblockA{\textit{Department of EECS} \\
\textit{York University}\\
Toronto, Canada \\
jihoshin@yorku.ca}
\and
\IEEEauthorblockN{2\textsuperscript{nd} Hadi Hemmati}
\IEEEauthorblockA{\textit{Department of EECS} \\
\textit{York University}\\
Toronto, Canada \\
hemmati@yorku.ca}
\and
\IEEEauthorblockN{3\textsuperscript{rd} Moshi Wei}
\IEEEauthorblockA{\textit{Department of EECS} \\
\textit{York University}\\
Toronto, Canada \\
moshiwei@yorku.ca}
\and
\IEEEauthorblockN{4\textsuperscript{th} Song Wang}
\IEEEauthorblockA{\textit{Department of EECS} \\
\textit{York University}\\
Toronto, Canada \\
wangsong@yorku.ca}
}

\maketitle
\begin{abstract}

Recently, deep learning models have shown promising results in test oracles generation. Static evaluation metrics from Natural Language Generation (NLG) such as BLEU, CodeBLEU, ROUGE-L, METEOR, and Accuracy, which is mainly based on textual comparisons, have been widely adopted to measure the performance of Neural Oracle Generation (NOG) models. However, these NLG-based metrics may not reflect the testing effectiveness of the generated oracle within a test suite, which is often measured by dynamic (execution-based) test adequacy metrics such as code coverage and mutation score.  


In this work, we revisit existing oracle generation studies plus \emph{ChatGPT} to empirically investigate the current standing of their performance in both NLG-based and test adequacy metrics. 
Specifically, we train and run four state-of-the-art test oracle generation models on five NLG-based and two test adequacy metrics for our analysis.   
We apply two different correlation analyses between these two different sets of metrics.
Surprisingly, we found no significant correlation between the NLG-based metrics and test adequacy metrics.
For instance, oracles generated from \emph{ChatGPT} on the project \emph{activemq-artemis} had the highest performance on all the NLG-based metrics among the studied NOGs, however, it had the most number of projects with a decrease in test adequacy metrics compared to all the studied NOGs.
We further conduct a qualitative analysis to explore the reasons behind our observations, we found that oracles with high NLG-based metrics but low test adequacy metrics tend to have complex or multiple chained method invocations within the oracle's parameters, making it hard for the model to generate completely, affecting the test adequacy metrics.
On the other hand, oracles with low NLG-based metrics but high test adequacy metrics tend to have to call different assertion types or a different method that functions similarly to the ones in the ground truth.
{Overall, this work complements prior studies on test oracle generation with an extensive performance evaluation with both NLG and test adequacy metrics and provides guidelines for better assessment of deep learning applications in software test generation in the future.}


\end{abstract}



\section{Introduction}
\label{sec:intro}



Unit testing is considered a standard procedure when developing a software product. 
The primary goal of unit testing is to verify whether a unit behaves as expected. However, writing an effective unit test takes a non-trivial effort for developers. To mitigate this challenge, there have been numerous studies to automate this process, the most recent of them exploiting deep neural generation models~\cite{liu2017automatic, saes2018unit, tufano2022methods2test}. 
While these models do not take into account which specific state (behavior) they want to verify (test oracle) and mainly try to catch higher-level exceptions or crash the software. 
Recently, researchers further propose test oracle generators based on deep learning and large language models, which aim to generate meaningful assertions and/or exception-handling logic for a method under test by providing the state they want to verify~\cite{watson2020learning, yu2022automated, dinella2022toga, tufano2022generating,yuan2023manual}. 

To evaluate neural oracle generation, metrics from natural language generation (NLG) such as BLEU \cite{papineni2002bleu}, CodeBLEU \cite{ren2020codebleu}, ROUGE-L \cite{lin2004rouge}, METEOR \cite{banerjee2005meteor}, and Accuracy \cite{accuracy1994} have been widely adopted to measure the performance of oracle generation models.
All these metrics are static and calculated based on the textual similarity of the generated oracle with a developer-written oracle in the training set. 
Although these generic metrics are useful in natural language applications, their benefit in special domains such as ours (test generation) is unclear. 
Traditionally, in software testing~\cite{fraser2011evosuite, pacheco2007randoop}, the quality of test case or test oracle is evaluated mainly based on test adequacy metrics, such as code coverage (i.e., the ability to explore code) and mutation score (i.e., the ability to find bugs). These metrics are dynamic and require the execution of the unit case under study. Thus they are much more expensive to collect and less attractive for researchers in neural test generation to work with.
Given that the dynamic test adequacy metrics are what ultimately matters in this domain and not the static NLG metrics, it seems that the implicit assumption of most existing research is that there is a high correlation between the high NLG metric and high test adequacy, to justify the use of these surrogate metrics. However, there is no study to approve or reject this assumption. 
Consider the following example in Listing \ref{lst:motive}, which shows an input and output example of an oracle generated by \emph{ChatGPT}.  
When the model receives an aggregation of the test prefix and the focal method, it generates an appropriate oracle that will go into the \emph{"<AssertPlaceHolder>"}.
As we can see, the generated oracle and the ground truth oracle function are the same (i.e., check whether `in' is null or not) with different types of assertion, hence the same test adequacy with the ground truth.
However, the corresponding BLEU score for the generated oracle is a low score of 0.21.
This calls for a more thorough investigation of the use of NLG-based metrics and test adequacy metrics in evaluating oracle generation tasks and their correlation. 
\definecolor{dkgreen}{rgb}{0,0.6,0}
\definecolor{gray}{rgb}{0.5,0.5,0.5}
\definecolor{mauve}{rgb}{0.58,0,0.82}

\lstset{frame=tb,
  language=Java,
  aboveskip=3mm,
  belowskip=3mm,
  showstringspaces=false,
  columns=flexible,
  basicstyle={\scriptsize\ttfamily},
  numbers=none,
  numberstyle=\tiny\color{gray},
  keywordstyle=\color{blue},
  commentstyle=\color{dkgreen},
  stringstyle=\color{mauve},
  breaklines=true,
  breakatwhitespace=true,
  tabsize=3
}

\vspace{10pt}
\begin{minipage}{\linewidth}
\begin{lstlisting}[caption={An example of oracle generation and its input and output.},captionpos=b,label={lst:motive}]
// Test prefix
testFileManagerNoFile () {
    FileManager fileManager = FileManager.create();
    fileManager.addLocatorFile();
    try {
        InputStream in = fileManager.open(filenameNonExistent);
        closeInputStream(in);
        "<AssertPlaceHolder>";
    } catch(NotFoundException ex) {
    }
}
// FocalMethod
private void closeInputStream(InputStream in ) {
    try {
    if ( in != null ) in.close ( ) ;
    } catch ( Exception ex ) {
    }
}

// Ground truth oracle
assertNull("Found non-existant file: " + filenameNonExistent, in)

// Generated oracle
assertEquals(null, in)

\end{lstlisting}
\end{minipage}


Therefore, to study the gap between these two categories of evaluation metrics, we first investigate current oracle generation literature to find their performance in various NLG-based metrics, i.e. BLEU, CodeBLEU, ROUGE-L, METEOR, and Accuracy.
Then we check the effectiveness of the oracles by measuring the test adequacy metrics, i.e., line coverage and mutation score. 
Finally, we calculate the correlations between ten pairs of evaluation metrics (five static NLG metrics and two dynamic test adequacy metrics).

The results show that the correlation between these two evaluation metric categories is not significant, no matter what metric we use. What this means in practice is that there are cases where the static metrics show an oracle is textually similar to a developer-written oracle but semantically the two oracles are quite different (their line coverage and mutation scores are very different). Or vice versa, that is, the static metrics suggest the two oracles are syntactically very different, but semantically they are very similar (e.g., same line coverage and mutation score). 
This has a significant implication for this research community on the way the current literature assesses proposed techniques in this domain, which requires revisiting. 

We further delve into examples from mismatching cases where generated oracles are showing high performance using one metric and low performance using another one. We also provide justifications for why this is happening based on our observed samples. For instance, we find that models fail to generate test oracles with complex or multiple chains of method invocation. These kinds of cases lead the oracles to be textually similar but semantically different (high NLG but low test metrics). Also, we found that some of the generated oracles have perfect functionality but are different in textual similarity (low NLG but high test metric). These disagreements were found to be the root cause of the low correlation between the two metrics.

In summary, our main contributions are as follows:
\begin{itemize}
    \item We revisit the performance of three state-of-the-art neural oracle generation models and \emph{ChatGPT} on both static NLG metrics and dynamic test adequacy metrics.
    \item We show (using two different correlation tests, on ten pairs of metrics) that there is no significant correlation between the two categories of metrics under study, which have significant implications for how the research in this domain must be done, in the future.
    \item We provide a manual analysis of the findings and provide justifications on why there is a mismatch between static and dynamic metrics in this domain, based on our observations.
    \item We provide a new benchmark for evaluating and studying NOG models, with almost 30K executable methods under test and their corresponding metrics from 13 open-source projects.
    We also release the dataset and source code of our experiments to help other researchers replicate and extend our study\footnote{\url{https://anonymous.4open.science/r/assessing_ntog-E307/}}.
\end{itemize}

The rest of this paper is organized as follows.
Section~\ref{sec:background} presents the background of neural oracle generation. 
Section~\ref{sec:setup} shows the experimental setup. 
Section~\ref{sec:results} presents the evaluation results. 
Section~\ref{sec:threats} discusses the threats to the validity of our study. 
Section~\ref{sec:related} presents the related studies. 
Section~\ref{sec:conclusion} concludes this paper. 
\section{Background on Neural Oracle Generation}
\label{sec:background}
Using deep generative models to generate useful test cases that reveal bugs are still very challenging, let alone generating complete and sound source code.
As a mitigation step, researchers proposed methods to generate meaningful test oracles, i.e. assertions, rather than generating the whole unit test methods.
A unit test case can be divided into two parts: the test prefix and the oracle.
Test prefix drives the system into an interesting state that the developer wants to test on. 
Oracles are the conditions that should be met in the state that the test prefix has stated.
Oracles are usually expressed with assertion statements or exception-handling logic.
Assertion statements ensure that a program gives true to the assertion statement, failing the program if the statement does not meet.
Exception handling logic can also be used for an oracle as it throws an exception if an undesired condition is met, or else passes the test method.
Assertion generation aims to generate assertion statements that can reveal the correct condition for the corresponding state given as a test prefix and its focal method.
There have been multiple studies regarding this field.
\emph{ATLAS}~\cite{watson2020learning} was the first study to exploit a deep neural generative model for generating assertion oracles. 
Yu and Lou et al. \cite{yu2022automated} extended the former by integrating simple information retrieval techniques, i.e. Jaccard coefficient \cite{tanimoto1958elementary}, Overlap \cite{wiki2023overlap}, and Dice coefficient \cite{dice1945measures}.
\emph{TOGA}~\cite{dinella2022toga} exploits a unified transformer-based neural model to generate both exception handling and assertion for unit test case oracles.
These proposed methods perform significantly well, ranging from 17.66\% \cite{watson2020learning} to as high as 96\% \emph{TOGA} in accuracy, according to the published literature. {Recently, researchers have examined whether \emph{ChatGPT} can achieve a remarkable performance on unit test generation~\cite{yuan2023manual,10172800,schafer2023adaptive}. For example, Yuan et al.~\cite{yuan2023manual} found that with appropriate prompts, \emph{ChatGPT} can help generate more compilable tests with correct assertions.} 

To evaluate neural oracle generation, metrics from natural language generation (NLG) such as BLEU \cite{papineni2002bleu}, CodeBLEU \cite{ren2020codebleu}, ROUGE-L \cite{lin2004rouge}, METEOR \cite{banerjee2005meteor}, and accuracy \cite{accuracy1994} have been widely adopted to measure the performance of oracle generation models.
\section{Experimental Design}
\label{sec:setup}
In this section, we explain the neural oracle generation models under study, our dataset, the evaluation metrics, the correlation analysis used in the paper, our research questions, and their corresponding procedures.  


\subsection{Neural Oracle Generators (NOGs)}
\label{sec:nogs}
In this study, we select three state-of-the-art (SOTA) NOG models proposed in the last three years. The NOGs are selected based on the following criteria:
\begin{itemize} 
    \item Should be a neural oracle generation model proposing a novel approach for oracle generations.
    \item The replication package to generate oracle should be publicly available.
    \item The model should be proposed in recent years and should represent the SOTA w.r.t. having a high NLG-based evaluation score. 
\end{itemize}
As a result, three NOG models were selected from 13 candidate papers published on main Software Engineering venues (ICSE, FSE, ASE, ISSTA, AST, TSE, TOSEM, EMSE, IST, JCST, and arXiv)
We also included \emph{ChatGPT} as a SOTA for LLMs, which has not been applied on NOG at the time of conducting this study but is a very relevant fit. 
The four selected NOGs in this study are the following:
\begin{itemize}
\item \textbf{ATLAS} \cite{watson2020learning}: is the first study to apply a deep neural generation model on oracle generation. It exploits a sequence-to-sequence encoder-decoder (RNN) model to learn and automatically generate an oracle when the test prefix and the focal method is given as input.
\item \textbf{IR} \cite{yu2022automated}: extends the previous study \emph{ATLAS} by leveraging information retrieval techniques and integrating deep learning to enhance performance. They also expanded the evaluation by adding unknown vocabulary to make the oracle generation problem more challenging.
\item \textbf{TOGA} \cite{dinella2022toga}: proposed a unified transformer-based neural model for both exceptional and assertion oracles. They first identify if the test method should raise an exception or generate an assertion. If it decides to raise an exception, it generates a try-catch clause. If not, it generates candidate assertions with different parameters. Then an assertion oracle ranker decides the best candidate for generating the assertion.
\item \textbf{ChatGPT} \cite{chatgpt}: is a large language model (LLM) powered chatbot developed by OpenAI. \emph{ChatGPT} is one of the most widely used LLMs currently. We wanted to investigate the impact of LLM as it has shown great potential in generating source code for software engineering tasks~\cite{yuan2023manual,10172800}. 
\end{itemize}

\subsection{Dataset}
\label{sec:dataset}

All three NOG models have a common dataset for evaluation, \emph{ATLAS} proposed in \cite{watson2020learning}.
It would be ideal to reuse the data to directly compare our results to each paper previously reported. 
The dataset is a parallel corpus of test prefixes plus the focal method paired with the oracle. 
However, to calculate the test adequacy metrics, we have to execute each test method and its corresponding focal method.
To execute them, we need the package information of the projects that they were originally mined from.
In the replication package of the \emph{ATLAS} paper, they provide the resulting raw dataset of test-oracle pairs and the project lists they used for mining.
Since the script that constructs the original dataset was not publicly available and each instance was randomly shuffled, retrieving the package information is not possible. 
So, we replicate and re-implement the \emph{ATLAS} dataset by following the steps reported in their papers.

The steps to mine the \emph{ATLAS} dataset are as follows:
First, we clone the project list that is publicly available in the original \emph{ATLAS} paper.
We extract all methods with the $@Test$ annotation, which is inherently used by the JUnit framework for unit test cases. 
For retrieving the oracle, we parse the test methods and look for the specific invocation of assertion APIs, e.g.,  assertEquals, assertTrue, assertNull, etc.
To parse the corresponding focal method, we first extract all methods declared within the project. 
Then we apply a heuristic to iterate all the invoked methods within the test method.
The invoked methods are queried to the list of declared methods in the project.
Then we get the last matched method before the oracle statement, assuming it's the focal method. 
Note that if there is a method invoked within the assertion parameter, we take the method as a focal method instead.
Test methods with more than one assert line are filtered as \emph{ATLAS} only focuses on single-line assertion generation as well as duplicate instances.

After retrieving the test-oracle pairs, we split the train, valid, and test set at a project level so that instances in the same project are not in different splits to remove potential information leakage within the projects. We randomly shuffle the projects and assign them to the train-valid-test set with a ratio of 8:1:1, which is the same ratio as the \emph{ATLAS} dataset. 
For evaluating projects on an execution level, we use a subset of the test set comprised of 13 projects as all the projects weren't deployable.
We considered the raw version of \emph{ATLAS} where we keep identifier names since it is a more challenging problem and \emph{IR} and \emph{TOGA} only consider them as well.
We also adapt the ideas of \emph{IR}'s new dataset that considers unknown vocabularies to better reflect the data distribution of the real world \cite{yu2022automated}.
We didn't consider \emph{Methods2Test} used in \emph{TOGA} as we only focus on assertion oracles rather than exceptional oracles or whole unit test case generations.
As shown in Table~\ref{tab:exp}, after re-implementing the data construction we resulted in a total of 323,994 test-oracle pairs, 261,794 for the train set, 32,476 for validation, and 29,724 for the test set.

\begin{table}[t!]
\centering
\caption{Experiment data}
\label{tab:exp}
\begin{tabular}{c|c|c|c|}
\cline{2-4}
    & Training & Validation & Test  \\ \hline
\multicolumn{1}{|c|}{\#instance} & 261.7K   & 32.4K      & 29.7K \\ \hline
\end{tabular}
\end{table}

\subsection{Evaluation Metrics}
\label{sec:metrics}
\subsubsection{NLG-based Metrics}
Existing oracle generation methods used NLG evaluation metrics which were originally adopted from the natural language processing (NLP) field to assess the performance of their models. 
The three NOGs used BLEU and Accuracy to assess their models.
However, researchers pointed out that BLEU and Accuracy is not suitable to assess code generation models due to the distinct difference between natural language and source code.
For a more comprehensive experiment, we have also added CodeBLEU \cite{ren2020codebleu} which is widely used to assess code generation models.
Also, we have added ROUGE-L and METEOR as they are commonly used in other NLG fields \cite{leclair2019recommendations, stapleton2020human}.

\begin{enumerate}
    \item \textbf{BLEU} \cite{papineni2002bleu} is a widely used evaluation in the NLG field that assesses the quality of text that has been generated by the model. It is calculated by an n-gram precision which is the number of matching n-grams from the generated and the ground truth text.
    \item \textbf{Accuracy (abbr. ACC) } \cite{accuracy1994} is the number of true predictions over the total amount of instances. The generated is a true positive only if it is an equivalent text with the ground truth.
    \item \textbf{CodeBLEU (abbr. CB)} \cite{ren2020codebleu} is an evaluation metric specifically designed for code generation models. It is calculated by the combination of a weighted n-gram precision (token similarity), AST matching (syntax similarity), and data flow matching (semantic similarity).
    \item \textbf{ROUGE-L (abbr. RL)} \cite{lin2004rouge} is another metric used in NLG. Out of the different ROUGE values, we report the ROUGE-L value which exploits the Longest Common Sub-sequence (LCS) in evaluating the similarity of the texts.
     \item \textbf{METEOR (abbr. MT)} \cite{banerjee2005meteor}  is also commonly used to evaluate NLG models. The metric is calculated by the harmonic mean of unigram precision and recall but with a higher weight on recall.
\end{enumerate}

\subsubsection{Test Adequacy Metrics}
In the field of automatic test case generation methods, researchers used test adequacy metrics to assess the quality of the generated test cases. Following existing work, we use line coverage and mutation score in this work to measure the test adequacy~\cite{tikir2002efficient,pacheco2005eclat,wang2021automatic}. 
\begin{enumerate}
    \item \textbf{Line Coverage}: is a common code metric to measure the number of lines covered when a test case is executed. It shows the strength of the test case in regards to their level of exercising the source code, line-wise.
    \item \textbf{Mutation Score}: is a fault-based metric to measure the quality of the generated test case by measuring how it can detect seeded faults (code mutants), effectively. The ability to detect a mutant means that the tests are likely to be effective at catching real faults as well. 
\end{enumerate}

\subsection{Correlation Analysis}
\label{sec:correlation}
To measure the correlation between each pair of metrics (i.e. one NLG-based and one test adequacy metric), we have chosen two correlation analysis methods, namely Spearman's and Kendall's rank correlation analysis.
The reason we chose these two methods is that we need a statistical test that works on non-parametric distributions and former studies which conduct correlation analysis on evaluation metrics commonly use them \cite{lim2023large, roy2021reassessing, shimorina2021human, graham2015accurate, liu2014strategy}.
Spearman's and Kendall's rank correlation measures the correlation between two ranked variables. They are non-parametric measures of statistical dependence between two variables and work on non-normal distributions. Spearman's calculation is calculated based on deviations from the mean ranks, while Kendall's correlation is calculated based on concordant and discordant pairs of ranks.

\subsection{Research Questions}
\label{sec:rqs}
\noindent\textbf{RQ1: How do current NOGs perform based on NLG-based evaluation metrics?}
In the literature, NOGs are evaluated mostly with two metrics, i.e. accuracy and BLEU score. 
We added three other majorly used NLG-based metrics, i.e. CodeBLEU, ROUGE-L, and METEOR, for a more comprehensive empirical study. Thus this RQ will show a thorough comparison among SOTA NOG models regarding five different NLG-based metrics.  


\noindent\textbf{RQ2: How do current NOGs perform based on test adequacy evaluation metrics?}
RQ1 compares different baselines based on the quality of the generated oracles measured by the NGL-based metrics. RQ2 repeats this experiment with dynamic test adequacy metrics. Since the goal of a NOG is to create an oracle as close as possible to a developer-written oracle, we measure how close the line coverage and mutation score of the generated oracles are to the developer-written ones. Thus the main motivation behind this RQ is whether we see the same trends when comparing baselines as RQ1 or not.

\noindent\textbf{RQ3: What is the correlation between NGL-based and test adequacy metrics?}
Following up on RQ2, the main motivation of this RQ is to come up with a recommendation for the research community on which static NGL-based metrics (if any) can safely be used as a surrogate measure for expensive adequacy metrics when evaluating their NOGs. A high/low correlation between an NGL-based metric and any test adequacy metric will be used as evidence of whether the NGL-based metric should be used in future research in this domain or not.

\noindent\textbf{RQ4: What are the reasons behind the correlation in RQ3?}
To further analyze the root cause of correlations or lack thereof found in RQ3, we manually analyze samples with mismatched correlation coefficients. Looking at these examples, we want to find out why some of the generated oracles have high scores in one metric category but low in the other. The goal is to provide justifications, based on the observed patterns, for the recommendations given in RQ3. 
 
\subsection{Experiment Procedure}
\label{sec:procedure}

\noindent \textbf{Experiment Setting for RQ1}:
We assess the NLG-based evaluation metrics, i.e. BLEU, CodeBLEU, ROUGE-L, METEOR, and accuracy on the studied NOGs. 
To generate oracles from the existing NOGs, i.e.,  \emph{ATLAS}, \emph{IR}, and \emph{TOGA}, we download the publicly available replication package shared in the paper and follow their instruction to train and run these tools on our dataset.   
We also report the BLEU and accuracy scores reported in the original paper for comparison. 

For \emph{ChatGPT}, following existing work~\cite{yuan2023manual}, we feed a single basic query to suggest a single line oracle to be replaced in the oracle placeholder given the text prefix and its focal method. 
We used gpt-3.5 API for the model version as it was the latest model accessible to us at the time of our experiment. 

We expect there would be a noticeable gap in the NLG-based metrics between our evaluation and the originally reported evaluation as we have divided the train-valid-test splits so that instances from the same project would fall into the same split. 
This is to mitigate any possible project-specific information leak between the train and the evaluation sets. 

\noindent \textbf{Experiment Setting for RQ2}: To calculate the test adequacy metrics, we have to build and execute each project and run the test cases with the generated oracles injected into the project. 
Since the \emph{ATLAS} dataset only considers single-line oracles, there are a lot of test cases in the repository that are not our target.
So we construct a sub-project for each project that only has the target test cases (those with single-line oracles). 
Specifically, we first remove all the test cases that are not in our test set for each project.
After we get the sub-projects, we replace the generated oracles with the original oracles written by the developers. 
Since the generated oracles are not guaranteed to be executable, after the replacement, we check if the modified test case with the generated oracle is still executable on the project (note that the original developer-written tests are all executable, before replacing their oracles).
We exclude test cases if they become unexecutable after the oracle injection as we need all test suites to be passing to calculate the test adequacy metrics. 
The resulting numbers of injected oracles are organized in Table \ref{tab:inj_asserts}.
We use PIT~\cite{coles2016pit} to calculate the test adequacy metrics, i.e. line coverage and mutation score. 
We also calculate and report (as the original score plus/minus the difference after the replacement) the test adequacy metrics per sub-project within the dataset.  
\begin{table*}[t!]
\centering
\caption{The number of injected oracles generated by each NOGs. Ms denotes the number of test methods, Cs denotes the number of affected classes, and Ss denotes the number of affected sub-modules.}
\vspace{-0.1in}
\resizebox{0.6\textwidth}{!}{
\begin{tabular}{l|ccc|ccc|ccc|ccc|}
\cline{2-13}
 & \multicolumn{3}{c|}{\textbf{ATLAS}} & \multicolumn{3}{c|}{\textbf{IR}} & \multicolumn{3}{c|}{\textbf{TOGA}} & \multicolumn{3}{c|}{\textbf{ChatGPT}} \\ \hline
\multicolumn{1}{|l|}{\textbf{project name}} & \multicolumn{1}{c|}{\textbf{Ms}} & \multicolumn{1}{c|}{\textbf{Cs}} & \textbf{Ss} & \multicolumn{1}{c|}{\textbf{Ms}} & \multicolumn{1}{c|}{\textbf{Cs}} & \textbf{Ss} & \multicolumn{1}{c|}{\textbf{Ms}} & \multicolumn{1}{c|}{\textbf{Cs}} & \textbf{Ss} & \multicolumn{1}{c|}{\textbf{Ms}} & \multicolumn{1}{c|}{\textbf{Cs}} & \textbf{Ss} \\ \hline
\multicolumn{1}{|l|}{activemq-artemis} & \multicolumn{1}{c|}{379} & \multicolumn{1}{c|}{177} & 21 & \multicolumn{1}{c|}{489} & \multicolumn{1}{c|}{211} & 21 & \multicolumn{1}{c|}{66} & \multicolumn{1}{c|}{43} & 13 & \multicolumn{1}{c|}{472} & \multicolumn{1}{c|}{210} & 21 \\ \hline
\multicolumn{1}{|l|}{cayenne} & \multicolumn{1}{c|}{417} & \multicolumn{1}{c|}{204} & 13 & \multicolumn{1}{c|}{460} & \multicolumn{1}{c|}{224} & 13 & \multicolumn{1}{c|}{94} & \multicolumn{1}{c|}{65} & 4 & \multicolumn{1}{c|}{463} & \multicolumn{1}{c|}{217} & 21 \\ \hline
\multicolumn{1}{|l|}{cloudstack} & \multicolumn{1}{c|}{410} & \multicolumn{1}{c|}{198} & 41 & \multicolumn{1}{c|}{466} & \multicolumn{1}{c|}{223} & 43 & \multicolumn{1}{c|}{190} & \multicolumn{1}{c|}{80} & 31 & \multicolumn{1}{c|}{466} & \multicolumn{1}{c|}{218} & 43 \\ \hline
\multicolumn{1}{|l|}{cxf} & \multicolumn{1}{c|}{709} & \multicolumn{1}{c|}{291} & 60 & \multicolumn{1}{c|}{1091} & \multicolumn{1}{c|}{329} & 62 & \multicolumn{1}{c|}{66} & \multicolumn{1}{c|}{52} & 22 & \multicolumn{1}{c|}{690} & \multicolumn{1}{c|}{323} & 62 \\ \hline
\multicolumn{1}{|l|}{drill} & \multicolumn{1}{c|}{295} & \multicolumn{1}{c|}{144} & 39 & \multicolumn{1}{c|}{386} & \multicolumn{1}{c|}{165} & 40 & \multicolumn{1}{c|}{73} & \multicolumn{1}{c|}{49} & 17 & \multicolumn{1}{c|}{349} & \multicolumn{1}{c|}{164} & 40 \\ \hline
\multicolumn{1}{|l|}{hadoop} & \multicolumn{1}{c|}{537} & \multicolumn{1}{c|}{356} & 41 & \multicolumn{1}{c|}{667} & \multicolumn{1}{c|}{426} & 41 & \multicolumn{1}{c|}{91} & \multicolumn{1}{c|}{67} & 20 & \multicolumn{1}{c|}{172} & \multicolumn{1}{c|}{92} & 2 \\ \hline
\multicolumn{1}{|l|}{ignite} & \multicolumn{1}{c|}{573} & \multicolumn{1}{c|}{377} & 15 & \multicolumn{1}{c|}{214} & \multicolumn{1}{c|}{120} & 8 & \multicolumn{1}{c|}{158} & \multicolumn{1}{c|}{112} & 9 & \multicolumn{1}{c|}{655} & \multicolumn{1}{c|}{409} & 15 \\ \hline
\multicolumn{1}{|l|}{itext7} & \multicolumn{1}{c|}{520} & \multicolumn{1}{c|}{242} & 10 & \multicolumn{1}{c|}{606} & \multicolumn{1}{c|}{266} & 10 & \multicolumn{1}{c|}{61} & \multicolumn{1}{c|}{39} & 9 & \multicolumn{1}{c|}{586} & \multicolumn{1}{c|}{257} & 10 \\ \hline
\multicolumn{1}{|l|}{jackrabbit-oak} & \multicolumn{1}{c|}{907} & \multicolumn{1}{c|}{504} & 29 & \multicolumn{1}{c|}{1107} & \multicolumn{1}{c|}{581} & 29 & \multicolumn{1}{c|}{398} & \multicolumn{1}{c|}{246} & 21 & \multicolumn{1}{c|}{1111} & \multicolumn{1}{c|}{566} & 29 \\ \hline
\multicolumn{1}{|l|}{james-project} & \multicolumn{1}{c|}{252} & \multicolumn{1}{c|}{121} & 45 & \multicolumn{1}{c|}{325} & \multicolumn{1}{c|}{138} & 48 & \multicolumn{1}{c|}{4} & \multicolumn{1}{c|}{4} & 1 & \multicolumn{1}{c|}{282} & \multicolumn{1}{c|}{129} & 46 \\ \hline
\multicolumn{1}{|l|}{jena} & \multicolumn{1}{c|}{815} & \multicolumn{1}{c|}{273} & 24 & \multicolumn{1}{c|}{920} & \multicolumn{1}{c|}{293} & 24 & \multicolumn{1}{c|}{132} & \multicolumn{1}{c|}{49} & 12 & \multicolumn{1}{c|}{930} & \multicolumn{1}{c|}{293} & 24 \\ \hline
\multicolumn{1}{|l|}{nifi} & \multicolumn{1}{c|}{468} & \multicolumn{1}{c|}{289} & 118 & \multicolumn{1}{c|}{541} & \multicolumn{1}{c|}{338} & 128 & \multicolumn{1}{c|}{159} & \multicolumn{1}{c|}{115} & 64 & \multicolumn{1}{c|}{546} & \multicolumn{1}{c|}{328} & 128 \\ \hline
\multicolumn{1}{|l|}{openmrs-core} & \multicolumn{1}{c|}{507} & \multicolumn{1}{c|}{157} & 3 & \multicolumn{1}{c|}{540} & \multicolumn{1}{c|}{170} & 3 & \multicolumn{1}{c|}{215} & \multicolumn{1}{c|}{82} & 3 & \multicolumn{1}{c|}{391} & \multicolumn{1}{c|}{132} & 1 \\ \hline \hline
\multicolumn{1}{|l|}{total} & \multicolumn{1}{c|}{6789} & \multicolumn{1}{c|}{3333} & 459 & \multicolumn{1}{c|}{7812} & \multicolumn{1}{c|}{3484} & 470 & \multicolumn{1}{c|}{1707} & \multicolumn{1}{c|}{1003} & 226 & \multicolumn{1}{c|}{7113} & \multicolumn{1}{c|}{3338} & 442 \\ \hline
\end{tabular}}
\label{tab:inj_asserts}
\end{table*}

\noindent \textbf{Experiment Setting for RQ3}: We conduct a correlation analysis between the NGL-based metrics and the test adequacy metrics. 
We get both metrics at the project level and perform correlation analysis using Spearman's and Kendall's rank correlation coefficient as they are the most common analysis done by different previous metric analysis studies as discussed in Section \ref{sec:correlation}.

The two variables that are measured for rankings are one NLG-based metric versus one test adequacy metric. 
Note that for the adequacy metrics, similar to RQ2, we work with the differences (deltas) between the score of the developer-written oracle and the generated oracle. In other words, we expect a delta score of zero for a perfectly generated oracle. A higher value of $|$delta$|$ means the two oracles are semantically different. Therefore, we calculate the correlations between each NLG-based metric (discussed in Section \ref{sec:metrics}) and the normalized $|$delta$|$ of line coverage and mutation scores (normalized using min-max). 

Since we have five NLG-based metrics and two test adequacy metrics, we have a total of 10 test runs, per project and baseline. However, the number of samples (projects) per baseline is limited (only 13 projects) which has a negative impact on the statistical tests p-values. We report this individual baseline analysis in the supplementary material in the replication package. But in the paper, we aggregate the four baseline data and look at a pool of 52 (13 X 4) samples per distribution (i.e., each distribution consists of 52 sample values for a given metric). Then we report the Rho and p-values per correlation test when comparing two metrics (each with a distribution of 52 samples).

From the results, we want to see how many of the metrics have a significant correlation between them, i.e. p-values of less than 0.05, and $\rho$ or $\tau$ values of at least 0.5. We also expect the $\rho$ values to be negative since we have a negative correlation between NLG-based and test adequacy metrics (e.g., high BLEU means the two oracle are similar whereas low $|$delta(mutation)$|$ means the same).

\noindent \textbf{Experiment Setting for RQ4}: In this RQ, we select 104 random samples of generated oracles to understand what actually causes the correlations. We specifically are interested in cases where there is a high disagreement between the NLG-based metrics and the adequacy metrics.
We pick random 2 samples from each project generated by each NOGs where the models are given the same input, i.e., the same generation problem (2 X 13 X 4 = 104). 

While doing the manual analysis we look for any patterns of syntax or semantics of the test code or the oracle itself that may have caused the disagreement.

\section{Experimental Results}
\label{sec:results}
\subsection{RQ1: NLG-based Metrics}
\label{sec:rq1_results}

\begin{table}[t!]
\centering
\caption{The performance of NOGs with our newly processed dataset. The column denotes the score of BLEU, accuracy (ACC), CodeBLEU (CB), ROUGE-L (RL), and METEOR (MT), respectively. The bold number shows the best metric score.}
\vspace{-0.1in}
\begin{tabular}{|l|l|l|l|l|l|}
\hline
\textbf{NOGs} & \textbf{BLEU} & \textbf{ACC} & \textbf{CB} & \textbf{RL} & \textbf{MT} \\ \hline
\emph{ATLAS} & 32.88 & 2.15 & 12.83 & 22.21 & 41.98 \\ \hline
\emph{IR} & 34.35 & 5.28 & 21.29 & 21.99 & 40.11 \\ \hline
\emph{TOGA} & 12.78 & 9.73 &  12.79 & 12.3 & 12.73 \\ \hline
\emph{ChatGPT} & \textbf{49.65} & \textbf{11.35} & \textbf{26.63} & \textbf{44.13} & \textbf{51.26} \\ \hline
\end{tabular}
\label{tab:rq1_nogs}
\end{table}
\begin{table}[t!]
\centering
\caption{Performance of NOGs reported in the original papers.}
\vspace{-0.1in}
\begin{tabular}{|c|c|c|}
\hline
\textbf{NOGs} & \textbf{BLEU} & \textbf{ACC} \\ \hline
\emph{ATLAS} & 61.85 & 17.66 \\ \hline
\emph{IR Old} & \textbf{78.86} & 46.54  \\ \hline
\emph{IR New} & 60.92 & 42.20  \\ \hline
\emph{TOGA} & - & \textbf{69.00} \\ \hline
\end{tabular}
\label{tab:rq1_original}
\end{table}

\begin{table}[t!]
\caption{The result of  NLG-based metrics on \emph{ATLAS}.}
\resizebox{0.8\columnwidth}{!}{
\begin{tabular}{|l|l|l|l|l|l|}
\hline
\textbf{Project Name} & \textbf{BLEU} & \textbf{ACC} & \textbf{CB} & \textbf{RL} & \textbf{MT} \\ \hline
activemq-artemis & 34.32 & 3.66 & 13.85 & 20.94 & 42.41 \\ \hline
cayenne & \textbf{40.66} & 0.72 & 17.76 & 31.24 & \textbf{50.42} \\ \hline
cloudstack & 30.42 & 0.24 & 11.43 & 26.22 & 46.51 \\ \hline
cxf & 34.20 & 0.14 & 10.19 & 25.05 & 47.17 \\ \hline
drill & 32.95 & 0.00 & 10.92 & 20.20 & 39.73 \\ \hline
hadoop & 31.85 & 0.74 & 12.26 & 22.39 & 41.51 \\ \hline
ignite & 36.81 & 1.05 & 18.29 & 28.25 & 46.91 \\ \hline
itext7 & 33.52 & \textbf{8.46} & 13.34 & 22.90 & 41.65 \\ \hline
jackrabbit-oak & 30.40 & 0.33 & 13.25 & 17.06 & 41.84 \\ \hline
james-project & 26.29 & 0.00 & \textbf{20.61} & 9.35 & 27.53 \\ \hline
jena & 35.46 & 0.86 & 7.61 & \textbf{32.67} & 47.28 \\ \hline
nifi & 32.87 & 2.35 & 14.72 & 28.20 & 49.86 \\ \hline
openmrs-core & 30.11 & 0.99 & 8.66 & 14.20 & 40.36 \\ \hline \hline
average & 33.07 & 1.50 & 13.30 & 22.97 & 43.32 \\ \hline
stdev & 3.54 & 2.33 & 3.81 & 6.72 & 5.96 \\ \hline
\end{tabular}
\label{tab:rq1_atlas}}
\end{table}
\begin{table}[t!]
\centering
\caption{The result of  NLG-based  metrics on \emph{IR}.}
\vspace{-0.1in}
\resizebox{0.8\columnwidth}{!}{
\begin{tabular}{|l|l|l|l|l|l|}
\hline
\textbf{Project Name} & \textbf{BLEU} & \textbf{ACC} & \textbf{CB} & \textbf{RL} & \textbf{MT} \\ \hline
activemq-artemis & 28.40 & 3.89 & 13.16 & 18.34 & 35.86 \\ \hline
cayenne & 35.89 & 0.00 & 11.72 & 34.17 & 41.24 \\ \hline
cloudstack & 44.83 & 18.24 & 27.18 & 34.17 & 48.51 \\ \hline
cxf & 32.71 & 1.19 & 14.12 & 19.66 & 38.44 \\ \hline
drill & 25.17 & 0.00 & 11.08 & 14.89 & 29.26 \\ \hline
hadoop & \textbf{53.08} & \textbf{29.54} & \textbf{33.97} & \textbf{44.37} & \textbf{55.77} \\ \hline
ignite & 26.61 & 0.80 & 8.68 & 13.52 & 33.15 \\ \hline
itext7 & 31.79 & 8.42 & 18.47 & 19.73 & 34.87 \\ \hline
jackrabbit-oak & 32.64 & 4.52 & 18.11 & 18.90 & 40.11 \\ \hline
james-project & 24.83 & 0.31 & 14.17 & 9.17 & 29.15 \\ \hline
jena & 34.00 & 6.65 & 16.28 & 38.48 & 50.59 \\ \hline
nifi & 34.00 & 6.65 & 16.28 & 22.10 & 41.20 \\ \hline
openmrs-core & 28.64 & 0.74 & 12.08 & 10.05 & 35.94 \\ \hline \hline
average & 33.28 & 6.23 & 16.56 & 22.89 & 39.55 \\ \hline
stdev & 7.98 & 8.65 & 6.96 & 11.26 & 8.04 \\ \hline
\end{tabular}}
\label{tab:rq1_ir}
\end{table}
\begin{table}[t!]
\centering
\caption{The result of  NLG-based metrics on \emph{TOGA}. }
\vspace{-0.1in}
\resizebox{0.8\columnwidth}{!}{
\begin{tabular}{|l|l|l|l|l|l|}
\hline
\textbf{Project Name} & \textbf{BLEU} & \textbf{ACC} & \textbf{CB} & \textbf{RL} & \textbf{MT} \\ \hline
activemq-artemis & 9.16 & 7.36 & 9.18 & 8.88 & 9.14 \\ \hline
cayenne & 11.70 & 5.70 & 8.27 & 10.63 & 11.51 \\ \hline
cloudstack & \textbf{25.81} & \textbf{23.97} & 19.34 & \textbf{25.42} & \textbf{25.74} \\ \hline
cxf & 5.78 & 5.38 & 5.79 & 5.74 & 5.78 \\ \hline
drill & 6.38 & 5.45 & 4.71 & 6.27 & 6.34 \\ \hline
hadoop & 8.04 & 6.92 & 8.02 & 7.79 & 7.98 \\ \hline
ignite & 16.92 & 11.57 & 17.13 & 16.06 & 16.91 \\ \hline
itext7 & 5.17 & 4.33 & 5.19 & 5.03 & 5.15 \\ \hline
jackrabbit-oak & 20.72 & 16.01 & \textbf{20.78} & 20.12 & 20.65 \\ \hline
james-project & 0.27 & 0.27 & 0.20 & 0.27 & 0.26 \\ \hline
jena & 8.55 & 7.08 & 8.54 & 8.35 & 8.54 \\ \hline
nifi & 17.04 & 12.73 & 12.56 & 16.26 & 16.26 \\ \hline
openmrs-core & 17.11 & 8.19 & 17.50 & 15.60 & 17.00 \\ \hline \hline
average & 11.74 & 8.84 & 10.55 & 11.26 & 11.64 \\ \hline
stdev & 7.26 & 6.05 & 6.37 & 7.00 & 7.19 \\ \hline
\end{tabular}}
\label{tab:rq1_toga}
\end{table}
\begin{table}[t!]
\centering
\caption{The result of  NLG-based metrics on \emph{ChatGPT}. The bold number shows the best metric score. The asterisk denotes the best metric score across all NOGs, when looking at the average scores.}

\resizebox{0.9\columnwidth}{!}{
\begin{tabular}{|l|l|l|l|l|l|}
\hline
\textbf{Project Name} & \textbf{BLEU} & \textbf{ACC} & \textbf{CB} & \textbf{RL} & \textbf{MT} \\ \hline
activemq-artemis & 45.11 & 9.32 & 28.66 & 40.48 & 48.78 \\ \hline
cayenne & 60.31 & 18.79 & \textbf{54.67} & 56.54 & 62.39 \\ \hline
cloudstack & 55.30 & 18.45 & 33.36 & 49.03 & 57.09 \\ \hline
cxf & 53.28 & 11.16 & 31.85 & 51.39 & 59.66 \\ \hline
drill & 47.36 & 6.30 & 35.58 & 39.67 & 45.30 \\ \hline
hadoop & 59.23 & 26.24 & 40.36 & 48.23 & 55.84 \\ \hline
ignite & 43.28 & 7.79 & 32.21 & 39.08 & 46.96 \\ \hline
itext7 & 49.78 & 17.58 & 25.80 & 46.24 & 51.49 \\ \hline
jackrabbit-oak & 44.78 & 7.29 & 26.04 & 37.64 & 49.13 \\ \hline
james-project & 43.68 & 3.19 & 26.04 & 31.78 & 34.60 \\ \hline
jena & \textbf{69.96} & \textbf{47.85} & 42.54 & \textbf{68.07} & \textbf{70.76} \\ \hline
nifi & 54.03 & 19.05 & 33.93 & 49.83 & 58.72 \\ \hline
openmrs-core & 58.91 & 10.79 & 29.44 & 48.00 & 56.61 \\ \hline \hline
average & \textbf{52.69*} & \textbf{15.68*} & \textbf{33.88*} & \textbf{46.61*} & \textbf{53.64*} \\ \hline
stdev & 8.03 & 11.69 & 8.14 & 9.34 & 9.05 \\ \hline
\end{tabular}}
\label{tab:rq1_chatgpt}
\end{table}
The results of the four studied NOGs are organized in Table \ref{tab:rq1_nogs}.
We show the scores of all five NLG-based metrics discussed in Section \ref{sec:metrics}.
We also organize the results reported in the original papers for comparison, in Table \ref{tab:rq1_original}.
In Table \ref{tab:rq1_original}, the results of \emph{ATLAS} are from evaluating the raw dataset (without normalizing identifier names) which is the version used by the other NOGs, i.e. \emph{IR} and \emph{TOGA}.
For the results of \emph{IR}, we report both the results from the old and new datasets.
\emph{IR Old} is the result of using the same dataset as \emph{ATLAS}, without unknown tokens.
\emph{IR New} is the new dataset that keeps instances that were dropped from the old dataset due to the unknown tokens.
For \emph{TOGA}, we report the results that used \emph{ATLAS} dataset for their assertion oracle generation.
We can see that \emph{TOGA} does not report the BLEU score as the methodology treats this problem as a ranking problem.
However, considering the first-ranked candidate assertion as a generation output, we can calculate the BLEU score.
So for comparison, we have included them in our results.

Using the original \emph{ATLAS} dataset, the best accuracy reported in the original papers is \emph{TOGA}. Since TOGA has a very high score on accuracy, it is very likely that it also achieves the best BLEU score because accuracy is a much more strict evaluation metric to achieve.
However, since they do not report the exact BLEU score, it is uncertain.
So the NOG that has the best reported BLEU score is generated from \emph{IR Old}.

Comparing what is reported in the original papers in Table \ref{tab:rq1_original} and the ones we evaluate in Table \ref{tab:rq1_nogs}, we can see that the BLEU and accuracy scores reported from the original papers have a big gap from the results we got from the newly processed data.
As mentioned in Section \ref{sec:rqs}, one possible reason for this is that the splitting of train-valid-test splits at a project level is impacted by removing information leaked into the new test set.
By keeping the instances from the same project in each split, similar code structures and project-dependent information is removed from the training set, making it harder for the NOG models to generate project-specific oracles.

Overall, in terms of the average scores across the projects, \emph{ChatGPT} exhibits the best performance among all NLG-based metrics (See average values with an asterisk in Table \ref{tab:rq1_chatgpt}). 
In Tables \ref{tab:rq1_atlas}-\ref{tab:rq1_chatgpt}, we also report the NLG-based metrics at a project level to investigate their effectiveness in each project.
Each table has a bold score for each metric score that achieves the highest with its model. For instance, \emph{ATLAS} achieves the highest BLEU and METEOR scores when being evaluated on the \emph{cayenne} project, the highest accuracy on \emph{itext7}, the highest CodeBLEU on \emph{james-project}, and the highest ROUGE-L score on \emph{jena}. For \emph{IR}, all the best score was achieved when evaluating the project \emph{hadoop}. For \emph{TOGA}, it achieves the highest onfour metrics evaluating \emph{cloudstack}, except for CodeBLEU, which is from \emph{jackrabbit-oak}. And lastly, \emph{ChatGPT} achieves the best on \emph{jena} for four metrics, except for CodeBLEU which achieves on the \emph{cayenne} project. 
Looking at all these project-level data, \emph{ChatGPT} has the highest per-project metric values as well (marked with an asterisk beside the scores in Table \ref{tab:rq1_chatgpt}).

Another clear observation is that the results of the baseline NOGs can vary a lot depending on the project and the variation is not consistent across the techniques. To summarize these variations we also report the standard deviation per baseline metric over the 13 projects. The results show that \emph{ChatGPT} had the highest standard deviation in four out of five metrics (only the \emph{IR} method's RL has a higher standard deviation than \emph{ChatGPT}'s).
Despite having the best performance concerning the NLG-based metrics, it was shown that it didn't have the best reliability in generating a consistent performance across different projects.


\mybox{From what is originally reported, \emph{TOGA} has the best overall score for NLG-based metrics. From evaluating the newly curated dataset, \emph{ChatGPT} has the best scores for all five NLG-based metrics in this experiment. There is a big drop in NLG-based metrics when we consider project-level for splitting. However, we also found that \emph{ChatGPT} had the highest standard deviation, showing that it has a less reliable performance in generating oracles in different projects.}

\begin{table*}[t!]
\centering
\caption{The result of test adequacy metrics for each NOGs. LC denotes the line coverage metric and MS denotes the mutation score. Column ±\% denotes the difference of absolute percentage points. The green color shows an increase after injection and the red shows a decrease after the injection.}

\resizebox{0.7\textwidth}{!}{
\begin{tabular}{l|llll|llll|llll|llll|}
\cline{2-17}
 & \multicolumn{4}{c|}{\textbf{ATLAS}} & \multicolumn{4}{c|}{\textbf{IR}} & \multicolumn{4}{c|}{\textbf{TOGA}} & \multicolumn{4}{c|}{\textbf{ChatGPT}} \\ \hline
\multicolumn{1}{|l|}{project name} & \multicolumn{1}{l|}{LC} & \multicolumn{1}{l|}{±\%} & \multicolumn{1}{l|}{MS} & ±\% & \multicolumn{1}{l|}{LC} & \multicolumn{1}{l|}{±\%} & \multicolumn{1}{l|}{MS} & ±\% & \multicolumn{1}{l|}{LC} & \multicolumn{1}{l|}{±\%} & \multicolumn{1}{l|}{MS} & ±\% & \multicolumn{1}{l|}{LS} & \multicolumn{1}{l|}{±\%} & \multicolumn{1}{l|}{MS} & ±\% \\ \hline
\multicolumn{1}{|l|}{activemq-artemis} & \multicolumn{1}{l|}{28.31} & \multicolumn{1}{l|}{\cellcolor[HTML]{FFCCC9}-0.02} & \multicolumn{1}{l|}{27.09} & \cellcolor[HTML]{9AFF99}0.79 & \multicolumn{1}{l|}{27.97} & \multicolumn{1}{l|}{\cellcolor[HTML]{FFCCC9}-0.23} & \multicolumn{1}{l|}{29.69} & \cellcolor[HTML]{FFCCC9}-0.26 & \multicolumn{1}{l|}{28.58} & \multicolumn{1}{l|}{\cellcolor[HTML]{9AFF99}\textbf{0.24}} & \multicolumn{1}{l|}{28.77} & {\cellcolor[HTML]{9AFF99}\textbf{1.65}} & \multicolumn{1}{l|}{22.71} & \multicolumn{1}{l|}{\cellcolor[HTML]{FFCCC9}-0.19} & \multicolumn{1}{l|}{21.68} & \cellcolor[HTML]{FFCCC9}-0.20 \\ \hline
\multicolumn{1}{|l|}{cayenne} & \multicolumn{1}{l|}{46.69} & \multicolumn{1}{l|}{0.00} & \multicolumn{1}{l|}{38.47} & \cellcolor[HTML]{9AFF99}0.11 & \multicolumn{1}{l|}{44.32} & \multicolumn{1}{l|}{0.00} & \multicolumn{1}{l|}{40.53} & \cellcolor[HTML]{FFCCC9}-0.49 & \multicolumn{1}{l|}{56.42} & \multicolumn{1}{l|}{0.00} & \multicolumn{1}{l|}{31.59} & \cellcolor[HTML]{9AFF99}0.56 & \multicolumn{1}{l|}{45.71} & \multicolumn{1}{l|}{0.00} & \multicolumn{1}{l|}{41.73} & \cellcolor[HTML]{9AFF99}0.32 \\ \hline
\multicolumn{1}{|l|}{cloudstack} & \multicolumn{1}{l|}{35.76} & \multicolumn{1}{l|}{0.00} & \multicolumn{1}{l|}{34.21} & \cellcolor[HTML]{FFCCC9}-0.15 & \multicolumn{1}{l|}{35.19} & \multicolumn{1}{l|}{0.00} & \multicolumn{1}{l|}{32.93} & \cellcolor[HTML]{FFCCC9}-0.20 & \multicolumn{1}{l|}{38.63} & \multicolumn{1}{l|}{0.00} & \multicolumn{1}{l|}{39.18} & \cellcolor[HTML]{FFCCC9}-0.40 & \multicolumn{1}{l|}{34.39} & \multicolumn{1}{l|}{\cellcolor[HTML]{9AFF99}1.73} & \multicolumn{1}{l|}{33.50} & \cellcolor[HTML]{9AFF99}1.88 \\ \hline
\multicolumn{1}{|l|}{cxf} & \multicolumn{1}{l|}{29.56} & \multicolumn{1}{l|}{0.00} & \multicolumn{1}{l|}{25.08} & 0.00 & \multicolumn{1}{l|}{28.60} & \multicolumn{1}{l|}{0.00} & \multicolumn{1}{l|}{24.75} & 0.00 & \multicolumn{1}{l|}{22.30} & \multicolumn{1}{l|}{0.00} & \multicolumn{1}{l|}{17.72} & \cellcolor[HTML]{9AFF99}0.24 & \multicolumn{1}{l|}{33.44} & \multicolumn{1}{l|}{\cellcolor[HTML]{9AFF99}\textbf{2.77}} & \multicolumn{1}{l|}{32.25} & \cellcolor[HTML]{9AFF99}0.89 \\ \hline
\multicolumn{1}{|l|}{drill} & \multicolumn{1}{l|}{49.04} & \multicolumn{1}{l|}{\cellcolor[HTML]{9AFF99}\textbf{0.41}} & \multicolumn{1}{l|}{36.68} & \cellcolor[HTML]{9AFF99}\textbf{1.06} & \multicolumn{1}{l|}{35.59} & \multicolumn{1}{l|}{0.00} & \multicolumn{1}{l|}{25.17} & \cellcolor[HTML]{9AFF99}\textbf{0.36} & \multicolumn{1}{l|}{14.64} & \multicolumn{1}{l|}{0.00} & \multicolumn{1}{l|}{5.49} & 0.00 & \multicolumn{1}{l|}{43.18} & \multicolumn{1}{l|}{\cellcolor[HTML]{9AFF99}2.63} & \multicolumn{1}{l|}{26.64} & \cellcolor[HTML]{9AFF99}2.30 \\ \hline
\multicolumn{1}{|l|}{hadoop} & \multicolumn{1}{l|}{21.02} & \multicolumn{1}{l|}{0.00} & \multicolumn{1}{l|}{13.33} & \cellcolor[HTML]{FFCCC9}-0.08 & \multicolumn{1}{l|}{27.31} & \multicolumn{1}{l|}{0.00} & \multicolumn{1}{l|}{23.53} & 0.00 & \multicolumn{1}{l|}{16.07} & \multicolumn{1}{l|}{0.00} & \multicolumn{1}{l|}{6.05} & 0.00 & \multicolumn{1}{l|}{21.02} & \multicolumn{1}{l|}{\cellcolor[HTML]{FFCCC9}-0.66} & \multicolumn{1}{l|}{17.23} & \cellcolor[HTML]{FFCCC9}-3.06 \\ \hline
\multicolumn{1}{|l|}{ignite} & \multicolumn{1}{l|}{48.93} & \multicolumn{1}{l|}{0.00} & \multicolumn{1}{l|}{42.95} & 0.00 & \multicolumn{1}{l|}{33.78} & \multicolumn{1}{l|}{0.00} & \multicolumn{1}{l|}{37.27} & 0.00 & \multicolumn{1}{l|}{22.86} & \multicolumn{1}{l|}{0.00} & \multicolumn{1}{l|}{25.90} & 0.00 & \multicolumn{1}{l|}{48.93} & \multicolumn{1}{l|}{0.00} & \multicolumn{1}{l|}{42.95} & \cellcolor[HTML]{FFCCC9}-0.72 \\ \hline
\multicolumn{1}{|l|}{itext7} & \multicolumn{1}{l|}{53.51} & \multicolumn{1}{l|}{0.00} & \multicolumn{1}{l|}{44.96} & \cellcolor[HTML]{FFCCC9}-0.41 & \multicolumn{1}{l|}{54.01} & \multicolumn{1}{l|}{0.00} & \multicolumn{1}{l|}{48.99} & \cellcolor[HTML]{FFCCC9}-0.15 & \multicolumn{1}{l|}{38.85} & \multicolumn{1}{l|}{0.00} & \multicolumn{1}{l|}{15.20} & 0.00 & \multicolumn{1}{l|}{51.96} & \multicolumn{1}{l|}{0.00} & \multicolumn{1}{l|}{46.13} & \cellcolor[HTML]{FFCCC9}-0.61 \\ \hline
\multicolumn{1}{|l|}{jackrabbit-oak} & \multicolumn{1}{l|}{36.10} & \multicolumn{1}{l|}{0.00} & \multicolumn{1}{l|}{33.93} & \cellcolor[HTML]{9AFF99}0.85 & \multicolumn{1}{l|}{36.76} & \multicolumn{1}{l|}{0.00} & \multicolumn{1}{l|}{42.67} & 0.00 & \multicolumn{1}{l|}{28.03} & \multicolumn{1}{l|}{0.00} & \multicolumn{1}{l|}{26.00} & \cellcolor[HTML]{9AFF99}0.06 & \multicolumn{1}{l|}{36.36} & \multicolumn{1}{l|}{\cellcolor[HTML]{FFCCC9}-0.68} & \multicolumn{1}{l|}{32.70} & \cellcolor[HTML]{FFCCC9}-0.83 \\ \hline
\multicolumn{1}{|l|}{james-project} & \multicolumn{1}{l|}{31.22} & \multicolumn{1}{l|}{0.00} & \multicolumn{1}{l|}{32.77} & \cellcolor[HTML]{FFCCC9}-2.87 & \multicolumn{1}{l|}{33.94} & \multicolumn{1}{l|}{\cellcolor[HTML]{FFCCC9}-2.74} & \multicolumn{1}{l|}{34.54} & \cellcolor[HTML]{FFCCC9}-1.73 & \multicolumn{1}{l|}{35.60} & \multicolumn{1}{l|}{0.00} & \multicolumn{1}{l|}{28.30} & 0.00 & \multicolumn{1}{l|}{33.94} & \multicolumn{1}{l|}{0.00} & \multicolumn{1}{l|}{33.07} & \cellcolor[HTML]{FFCCC9}-1.39 \\ \hline
\multicolumn{1}{|l|}{jena} & \multicolumn{1}{l|}{52.75} & \multicolumn{1}{l|}{\cellcolor[HTML]{FFCCC9}-0.05} & \multicolumn{1}{l|}{52.27} & \cellcolor[HTML]{9AFF99}0.76 & \multicolumn{1}{l|}{\textbf{59.94}} & \multicolumn{1}{l|}{0.00} & \multicolumn{1}{l|}{\textbf{60.23}} & 0.00 & \multicolumn{1}{l|}{38.62} & \multicolumn{1}{l|}{0.00} & \multicolumn{1}{l|}{35.43} & 0.00 & \multicolumn{1}{l|}{59.87} & \multicolumn{1}{l|}{\cellcolor[HTML]{9AFF99}0.52} & \multicolumn{1}{l|}{57.53} & \cellcolor[HTML]{9AFF99}\textbf{3.23} \\ \hline
\multicolumn{1}{|l|}{nifi} & \multicolumn{1}{l|}{\textbf{58.16}} & \multicolumn{1}{l|}{0.00} & \multicolumn{1}{l|}{\textbf{58.48}} & 0.00 & \multicolumn{1}{l|}{58.16} & \multicolumn{1}{l|}{0.00} & \multicolumn{1}{l|}{58.48} & 0.00 & \multicolumn{1}{l|}{\textbf{58.37}} & \multicolumn{1}{l|}{0.00} & \multicolumn{1}{l|}{\textbf{53.38}} & 0.00 & \multicolumn{1}{l|}{\textbf{63.41}} & \multicolumn{1}{l|}{0.00} & \multicolumn{1}{l|}{\textbf{59.74}} & \cellcolor[HTML]{FFCCC9}-0.46 \\ \hline
\multicolumn{1}{|l|}{openmrs-core} & \multicolumn{1}{l|}{8.25} & \multicolumn{1}{l|}{0.00} & \multicolumn{1}{l|}{4.82} & 0.00 & \multicolumn{1}{l|}{8.25} & \multicolumn{1}{l|}{0.00} & \multicolumn{1}{l|}{4.82} & 0.00 & \multicolumn{1}{l|}{8.25} & \multicolumn{1}{l|}{0.00} & \multicolumn{1}{l|}{4.82} & 0.00 & \multicolumn{1}{l|}{8.25} & \multicolumn{1}{l|}{0.00} & \multicolumn{1}{l|}{4.85} & \cellcolor[HTML]{FFCCC9}-0.03 \\ \hline \hline
\multicolumn{1}{|l|}{median} & \multicolumn{1}{l|}{36.10} & \multicolumn{1}{l|}{0.00} & \multicolumn{1}{l|}{34.21} & 0.00 & \multicolumn{1}{l|}{35.19} & \multicolumn{1}{l|}{0.00} & \multicolumn{1}{l|}{34.54} & 0.00 & \multicolumn{1}{l|}{28.58} & \multicolumn{1}{l|}{0.00} & \multicolumn{1}{l|}{26.00} & 0.00 & \multicolumn{1}{l|}{36.36} & \multicolumn{1}{l|}{0.00} & \multicolumn{1}{l|}{33.07} & \cellcolor[HTML]{FFCCC9}-0.20 \\ \hline
\multicolumn{1}{|l|}{average} & \multicolumn{1}{l|}{38.41} & \multicolumn{1}{l|}{\cellcolor[HTML]{9AFF99}0.03} & \multicolumn{1}{l|}{34.23} & 0.00 & \multicolumn{1}{l|}{37.22} & \multicolumn{1}{l|}{\cellcolor[HTML]{FFCCC9}-0.23} & \multicolumn{1}{l|}{35.66} & \cellcolor[HTML]{FFCCC9}-0.19 & \multicolumn{1}{l|}{31.32} & \multicolumn{1}{l|}{\cellcolor[HTML]{9AFF99}0.02} & \multicolumn{1}{l|}{24.45} & \cellcolor[HTML]{9AFF99}0.16 & \multicolumn{1}{l|}{38.71} & \multicolumn{1}{l|}{\cellcolor[HTML]{9AFF99}0.47} & \multicolumn{1}{l|}{34.62} & \cellcolor[HTML]{9AFF99}0.10 \\ \hline
\multicolumn{1}{|l|}{stdev} & \multicolumn{1}{l|}{14.64} & \multicolumn{1}{l|}{0.12} & \multicolumn{1}{l|}{14.63} & 0.98 & \multicolumn{1}{l|}{14.18} & \multicolumn{1}{l|}{0.76} & \multicolumn{1}{l|}{15.14} & 0.50 & \multicolumn{1}{l|}{15.16} & \multicolumn{1}{l|}{0.07} & \multicolumn{1}{l|}{14.37} & 0.49 & \multicolumn{1}{l|}{15.75} & \multicolumn{1}{l|}{1.15} & \multicolumn{1}{l|}{15.39} & 1.66 \\ \hline
\end{tabular}}
\label{tab:rq2}
\end{table*}

\subsection{RQ2: Test Adequacy Metrics}
\label{sec:rq2_results}
The overall results of test adequacy metrics are shown in Table \ref{tab:rq2}.
We also report the box plot of test adequacy metrics to compare the different NOGs evaluated in this study.
As discussed in Section \ref{sec:procedure}, we execute the sub-projects before and after replacing the oracles generated by each studied NOGs.
The test adequacy metrics scores reported in Table \ref{tab:rq2} are calculated before the injection and the column denoted with ±\% on the right side of each metric shows the absolute percentage point difference by calculating the metrics after the generated oracles are injected.

As explained in Section \ref{sec:procedure}, the level of increase or decrease in the test adequacy metrics is not big as we are expecting the generated oracle to be similar to the developer-written one.
For the results of \emph{ATLAS}, we can see that the highest increase in line coverage and mutation score is from evaluating the \emph{drill} project. For \emph{IR}, \emph{drill} had the highest increase in mutation score but no projects were increased for line coverage. For \emph{TOGA}, \emph{activemq-artemis} had the highest increase in both line coverage and mutation score. For \emph{ChatGPT}, \emph{cxf} had the highest increase in line coverage and \emph{jena} in mutation score.

To compare which NOG has the best performance overall, we can observe that \emph{ChatGPT} has the highest increase in line coverage when evaluating \emph{cxf} and mutation score when evaluating \emph{jena}. If we consider the average increase of line coverage, \emph{ChatGPT} still has the best performance while \emph{TOGA} has the best performance in the average increase of mutation score. Overall, \emph{ChatGPT} has the highest performance which is in line with the results from RQ1. However, it is also very interesting to find that \emph{ChatGPT} also has the most number of projects that have a decrease in the test adequacy metrics with 11 out of 26 fields, which accounts for around 42\%.
This could also tell us that despite having a very high score in NLG-based metrics, it could have a negative impact on the quality or the strength of the generated oracle. This calls to our attention that we must be careful in using NLG-based metrics as the standard to evaluate NOGs.

Also, unlike what we have observed from NLG-based metrics in RQ1, there was a similar trend in the ranges of scores in test adequacy metrics for projects throughout the NOGs.
For instance, the metric scores for \emph{activemq-artemis} ranged in the 20s, \emph{cloudstack} ranged in the 30s, \emph{james-project} ranged in the 30s, etc.
Some projects had higher variance than others, but the patterns and ranges are much clearer than the ones we observed in the results for NLG-based metrics. 
One reason that we suspect is that the NLG-based metric has a high fluctuation
in the score if the model generates even one token that is different from the ground truth text.
Since the task is to generate a one-liner assertion oracle with shorter tokens, the metric will have a very big decrease.
However for test adequacy, even though the generation is different token-wise, if it correctly generates the same type and values in a similar range for an argument, the executed line of code will be similar.

Additionally, \emph{ChatGPT} showed the highest standard deviation, showing unreliable performance for test adequacy metrics.

\mybox{Overall, \emph{ChatGPT} showed the best performance in the test adequacy metrics concerning the maximum score it reaches and the average score. However, we also found that \emph{ChatGPT} had the most number of projects that have decreased in the test adequacy metrics and the highest standard deviation, showing us that it can be unreliable on different projects. We also found that test adequacy metrics are more stable and have a much clearer trend than NLG-based metrics based on the variation of scores across different projects.}

\begin{figure}[t!]
\centering
\resizebox{0.8\columnwidth}{!}{
\begin{subfigure}{0.45\columnwidth}
\scalebox{0.5}{
\begin{tikzpicture}
\pgfplotsset{
	   boxplot/every whisker/.style={thick,solid,black},
	   boxplot/every median/.style={very thick,solid,black},
}
\begin{axis}[
xtick={1,2,3,4},
xticklabels= ,
boxplot/draw direction=y,
ymin= 0,
ymax= 0.7,
]

\addplot[
very thick,
mark=+,
fill=white,
boxplot
]
table [row sep=\\, y index=0] {
	data\\0.2829\\0.4669\\0.3576\\0.2956\\0.4944\\0.2102\\0.4893\\0.5351\\0.3610\\0.3122\\0.5270\\0.5816\\0.0825\\};

\addplot[
very thick,
mark=+,
fill=white,
boxplot
]
table [row sep=\\, y index=0] {
	data\\0.2773\\0.4432\\0.3519\\0.2860\\0.3559\\0.2731\\0.3378\\0.5401\\0.3676\\0.3120\\0.5994\\0.5816\\0.0825\\};

\addplot[
very thick,
mark=+,
fill=white,
boxplot
]
table [row sep=\\, y index=0] {
    data\\0.2883\\0.5642\\0.3863\\0.2230\\0.1464\\0.1607\\0.2286\\0.3885\\0.2803\\0.3560\\0.3862\\0.5837\\0.0825\\};

\addplot[
very thick,
mark=+,
fill=white,
boxplot
] 
table [row sep=\\, y index=0] {
    data\\0.2252\\0.4571\\0.3612\\0.3621\\0.4581\\0.2036\\0.4893\\0.5196\\0.3569\\0.3394\\0.6039\\0.6341\\0.0825\\};

\end{axis}
\end{tikzpicture}
}
\caption{\small Line coverage}
\end{subfigure}
\begin{subfigure}{0.45\columnwidth}
\scalebox{0.5}{
\begin{tikzpicture}
\pgfplotsset{
	   boxplot/every whisker/.style={thick,solid,black},
	   boxplot/every median/.style={very thick,solid,black},
}
\begin{axis}[
xtick={1,2,3,4},
xticklabels= ,
boxplot/draw direction=y,
ymin= 0,
ymax= 0.65,
]

\addplot[
very thick,
mark=+,
fill=white,
boxplot
]
table [row sep=\\, y index=0] {
	data\\0.2788\\0.3858\\0.3406\\0.2508\\0.3774\\0.1325\\0.4295\\0.4455\\0.3478\\0.2990\\0.5303\\0.5848\\0.0482\\};

\addplot[
very thick,
mark=+,
fill=white,
boxplot
]
table [row sep=\\, y index=0] {
	data\\0.2943\\0.4005\\0.3273\\0.2475\\0.2553\\0.2353\\0.3727\\0.4884\\0.4267\\0.3281\\0.6023\\0.5848\\0.0482\\};

\addplot[
very thick,
mark=+,
fill=white,
boxplot
]
table [row sep=\\, y index=0] {
    data\\0.3042\\0.3215\\0.3878\\0.1796\\0.0549\\0.0605\\0.2590\\0.1520\\0.2606\\0.2830\\0.3543\\0.5338\\0.0482\\};

\addplot[
very thick,
mark=+,
fill=white,
boxplot
] 
table [row sep=\\, y index=0] {
    data\\0.2148\\0.4205\\0.3538\\0.3314\\0.2894\\0.1416\\0.4223\\0.4552\\0.3187\\0.3168\\0.6076\\0.5928\\0.0482\\};

\end{axis}
\end{tikzpicture}
}
\caption{\small Mutation score}
\end{subfigure}}

\caption{Line coverage and mutation score after oracle injection generated by the four NOGs. The order of boxes are \emph{ATLAS}, \emph{IR}, \emph{TOGA}, and \emph{ChatGPT} ,respectively.}
\label{fig:recalls}
\end{figure}
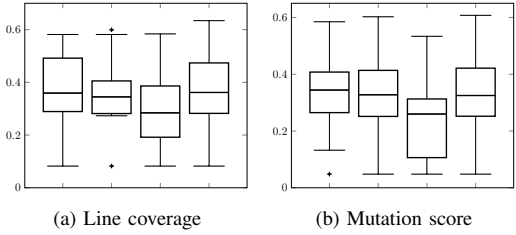

\subsection{RQ3: Correlation Analysis}
\label{sec:rq3_results}
We report the correlation coefficient and the p-value of both Spearman's and Kendall's rank correlation, in Table \ref{tab:rq3_correlation}.
The bold number in the table shows p-values less than 0.05 which shows the statistical significance of the results.
The different colors in the cell show the level of agreement that each coefficient has.
Red color shows a very weak agreement, orange color shows a weak agreement, and green color shows a moderate agreement.
For $\rho$, 0.01 to 0.10 is considered a very weak agreement, 0.10 to 0.39 is considered a weak agreement, 0.4 to 0.69 is considered a moderate correlation, 0.70 to 0.89 is considered a strong correlation, and 0.90 to 1.00 is considered a very strong correlation \cite{schober2018correlation}.
For $\tau$, 0.01 to 0.24 is considered a very weak agreement, 0.25 to 0.34 is considered a weak agreement, 0.35 to 0.39 is considered a moderate agreement, and 0.40 to 1.0 is considered a strong agreement~\cite{smith1981scope}.

As we can see from the results, except for LC and MS vs Accuracy, the tests from all other pairs result in p-values less than 0.05, which shows their correlation results are statistically significant.
However, all correlations are either very low or low, using either test. The highest correlation is for MS vs BLEU with $\rho$ = -0.39, which is still considered as low.  



Considering these results, we would recommend NOT using NGL-based metrics as the main evaluation metric when studying NOGs. Given that calculating test adequacy metrics is expensive (requires execution), they might not be feasible or cost-beneficial, especially during the training phase. Therefore, researchers may use BLEU (since it had the highest correlation among the five) as their initial metric but the final decisions must be made based on test adequacy metrics. 



\mybox{Using Spearman's and Kendall's rank correlation analysis, we found that the most closely correlated NLG-based metric to the test adequacy metrics was the BLEU score with $\rho(BLEU vs MS)$=-0.39, which is considered a weak correlation. Based on this finding we recommend NOG researchers avoid only using NLG-based metrics when evaluating NOGs and make sure to include at least one test adequacy metric as their main metric.}

\begin{table}[t!]
\centering
\caption{The result of correlation analysis on all the results from studied NOGs. The bold number shows the p-value that is less than 0.05 which shows the observed correlations are also statistically significant. Red cell shows ``Very Weak Agreement'' and orange cell shows ``Weak Agreement''.}

\resizebox{0.8\columnwidth}{!}{
\begin{tabular}{c|cc|cc|}
\cline{2-5}
\multicolumn{1}{l|}{} & \multicolumn{2}{c|}{\textbf{Spearman's}} & \multicolumn{2}{c|}{\textbf{Kendall's}} \\ \hline
\multicolumn{1}{|c|}{\textbf{Metrics}} & \multicolumn{1}{c|}{\textbf{rho}} & \textbf{P-value} & \multicolumn{1}{c|}{\textbf{tau}} & \textbf{P-value} \\ \hline
\multicolumn{1}{|c|}{LC vs BLEU} & \multicolumn{1}{c|}{\cellcolor[HTML]{FFCE93}-0.3461} & \textbf{0.0070} & \multicolumn{1}{c|}{\cellcolor[HTML]{FFCE93}-0.2760} & \textbf{0.0057} \\ \hline
\multicolumn{1}{|c|}{LC vs ACC} & \multicolumn{1}{c|}{\cellcolor[HTML]{FD6864}-0.0394} & 0.3944 & \multicolumn{1}{c|}{\cellcolor[HTML]{FD6864}-0.0342} & 0.3773 \\ \hline
\multicolumn{1}{|c|}{LC vs CB} & \multicolumn{1}{c|}{\cellcolor[HTML]{FFCE93}-0.2806} & \textbf{0.0250} & \multicolumn{1}{c|}{\cellcolor[HTML]{FD6864}-0.2225} & \textbf{0.0223} \\ \hline
\multicolumn{1}{|c|}{LC vs RL} & \multicolumn{1}{c|}{\cellcolor[HTML]{FFCE93}-0.3096} & \textbf{0.0148} & \multicolumn{1}{c|}{\cellcolor[HTML]{FFCE93}-0.2499} & \textbf{0.0110} \\ \hline
\multicolumn{1}{|c|}{LC vs MT} & \multicolumn{1}{c|}{\cellcolor[HTML]{FFCE93}-0.3347} & \textbf{0.0090} & \multicolumn{1}{c|}{\cellcolor[HTML]{FFCE93}-0.2600} & \textbf{0.0086} \\ \hline
\multicolumn{1}{|c|}{MS vs BLEU} & \multicolumn{1}{c|}{\cellcolor[HTML]{FFCE93}-0.3982} & \textbf{0.0021} & \multicolumn{1}{c|}{\cellcolor[HTML]{FFCE93}-0.2894} & \textbf{0.0019} \\ \hline
\multicolumn{1}{|c|}{MS vs ACC} & \multicolumn{1}{c|}{\cellcolor[HTML]{FD6864}-0.0927} & 0.2637 & \multicolumn{1}{c|}{\cellcolor[HTML]{FD6864}-0.0695} & 0.2444 \\ \hline
\multicolumn{1}{|c|}{MS vs CB} & \multicolumn{1}{c|}{\cellcolor[HTML]{FFCE93}-0.3198} & \textbf{0.0121} & \multicolumn{1}{c|}{\cellcolor[HTML]{FD6864}-0.2320} & \textbf{0.0114} \\ \hline
\multicolumn{1}{|c|}{MS vs RL} & \multicolumn{1}{c|}{\cellcolor[HTML]{FFCE93}-0.3286} & \textbf{0.0102} & \multicolumn{1}{c|}{\cellcolor[HTML]{FD6864}-0.2307} & \textbf{0.0106} \\ \hline
\multicolumn{1}{|c|}{MS vs MT} & \multicolumn{1}{c|}{\cellcolor[HTML]{FFCE93}-0.3178} & \textbf{0.0126} & \multicolumn{1}{c|}{\cellcolor[HTML]{FD6864}-0.2249} & \textbf{0.0124} \\ \hline
\end{tabular}}
\label{tab:rq3_correlation}
\end{table}

\subsection{RQ4: Manual Analysis}
\label{sec:rq4_results}
To further investigate why these two metrics are not significantly correlated, we manually inspect the randomly selected 104 examples of the generated oracles, from all projects.
We have found three major types of patterns in the generated oracles. 1) Type-1: identical, 2) Type-2: textually different but same/similar functions, 3) and Type-3: textually similar but different in semantics.
Type-1 exhibits a high NLG-based metric and a close to zero increase/decrease in test adequacy metric. That means both metrics agree that the generated oracle and the developer-written one are very similar. 
Type-2 and 3 are the interesting ones.
Type-2 exhibits a low score for NLG-based metrics, however, it achieves close to zero increase/decrease in test adequacy. This means the two oracles are textually different but semantically (from testing perspective) similar. 
Type-3 also shows a disagreement between the metrics by exhibiting high NLG-based metrics (similar textually) but creating in the test adequacy metrics. 

From the 104 samples we inspected, we found 36 Type-1 examples, 37 Type-2 examples, and 1 Type-3 example. \emph{TOGA} had the most Type-1 with 22, \emph{ChatGPT} had the most Type-2 with 17 and the only Type-3 as well.
We give an example code to demonstrate the disagreeing examples. As introduced in Section~\ref{sec:intro}, Listing \ref{lst:motive} shows a perfect example of Type-2, with a high textual difference but the same functions. Oracles with this type tend to have different assertion types to address checking the states or call different methods in their parameters that function similarly or identically beside the methods invoked in the ground truth oracle. Listing \ref{lst:rq4} is an example of Type-3 where it is textually similar but not semantically. This example has a close textual similarity where the parameters are all included in the ground truth oracle with the BLEU score of 0.54. However, the type of assertion is different and the generated oracle doesn't cover the lines for $IsoMatcher.isomorphic()$. This would affect the test adequacy metrics. The ground truth oracles of this type had multiple or complex method invocations within the parameter, making it hard for the model to generate. These examples show how the generated oracles can cause the two metrics to disagree with each other. 

\mybox{From our manual analysis, we have found three major types of patterns in the generated oracles, and two of them contribute to the nonalignment of NGL-based and test adequacy metrics. Type-1 is an oracle generated with a close agreement to ground truth (high NLG, high test metric), Type-2 is an oracle with disagreement by low NLG but high test metric, and lastly Type-3 with disagreement by high NLG but low test metric. We found Type-2 is the most common. They tend to have different methods for achieving a similar function. Type-3 wasn't found much however it was due to the model's failure to generate the whole sequence of method chain that hindered the test adequacy metric. The first two types were the cause of major disagreement between the two different metric sets.}

\definecolor{dkgreen}{rgb}{0,0.6,0}
\definecolor{gray}{rgb}{0.5,0.5,0.5}
\definecolor{mauve}{rgb}{0.58,0,0.82}

\lstset{frame=tb,
  language=Java,
  aboveskip=3mm,
  belowskip=3mm,
  showstringspaces=false,
  columns=flexible,
  basicstyle={\scriptsize\ttfamily},
  numbers=none,
  numberstyle=\tiny\color{gray},
  keywordstyle=\color{blue},
  commentstyle=\color{dkgreen},
  stringstyle=\color{mauve},
  breaklines=true,
  breakatwhitespace=true,
  tabsize=3
}

\vspace{10pt}
\begin{minipage}{\linewidth}
\begin{lstlisting}[caption={An example Type-3, disagreement by textually similar but not in test adequacy.},captionpos=b,label={lst:rq4}]
// Ground truth assertion
assertTrue(IsoMatcher.isomorphic(dsgData, dataset.asDatasetGraph()))

// Generated oracle
assertEquals(dsgData, dataset.asDatasetGraph())
\end{lstlisting}
\end{minipage}
\section{Threats to Validity}
\label{sec:threats}

\noindent \textbf{Internal Validity.} To avoid any confounding factors, we use metrics that are used in the literature for both static and dynamic metrics and use the same implementation of the models and the metrics from the original papers as much as possible.


\noindent \textbf{Construct Validity.} The main threat to the construct validity can be the evaluation metrics we used.
The test adequacy metrics we used in this study are line coverage and mutation score.
Although these metrics are widely used to assess the effectiveness of test cases, they may not be highly correlated to finding actual bugs (the ultimate goal of testing).
In our future study, we plan to examine correlations to real bug detection as well.

\noindent \textbf{Conclusion Validity.} We have conducted two statistical tests and carefully analyzed the statistical significance when reporting correlations, to avoid conclusion validity threats. 

\noindent \textbf{External Validity.} 
The main threats to external validity in this study are the limitations to (a) baseline models (b) test adequacy metrics, and (c) the dataset size. 
Regarding baselines, we have used the most recently published state-of-the-art performing models of neural oracle generation models that were publicly available and that proposed a novel way of generating assertion statement lines when the test prefix and focal method are given. Including other baselines might change the observations from this study and will be worth exploring when they are proposed in the future.
Regarding test adequacy metrics, we use the line coverage and mutation score. It's possible that our findings are not applicable to other levels of code coverage, i.e., branch coverage, statement coverage, or method coverage. In the future, we plan to include more types of code coverage.
Finally, regarding the dataset, although we have put much effort into maintaining high-quality data by using real-world test methods contributed by developers on GitHub and using multiple varieties in the domain of projects, it could not be enough to generalize to all data points in the world.
This was inevitable as we had limited resources to assess projects since we had to build, execute, and run the entire project to get test adequacy metrics which takes a lot of time and computational complexity.
However, we deem we have devised a good number of projects from different domains, and a good number of instances in test methods, to show significant observations from the used test set.
\section{Related Work}
\label{sec:related}

\subsection{Traditional Test Oracle Generation}
In Section \ref{sec:background}, we already discussed the related work on neural test oracle generation. However, there have been numerous test oracle generation techniques that do not use neural generation models. In this subsection we very briefly mention them. 
Peter and Parnas \cite{peters1994generating} proposed a test oracle generator tool that uses relational program specifications or documents to generate expected outputs of tests as tabular expressions.
Bousqpent et al. \cite{du1999lutess} proposed \emph{Lutess}, a framework that automatically constructs test harnesses from various formal descriptions, i.e. software environment constraints, functional and safety-oriented properties, software operation profiles, and software behavior patterns. 
Shahamiri et al. \cite{shahamiri2011automated} proposed an automated test oracle framework using I/O relational analysis to generate the output domain, multi-networks oracle for input-to-output domain mapping, and a comparator to adjust the precision of generated oracle by defining the comparison tolerance.
Liu and Shin \cite{liu2020automatic} proposed a new method, \emph{V-method} for automatic test case and test oracle generation from model-based formal specifications. They exploit functional scenarios defined in the formal specification, test generation criteria, algorithms, and mechanisms for deriving test oracles.

\subsection{Re-evaluating Evaluation Metrics}


Since our study is about re-evaluating NOG evaluation metrics, in this subsection, we also cover most related work that re-evaluates evaluation metrics but in other domains than NOG. 
Recently, there have been numerous studies about revisiting evaluation metrics in the NLG field.
Reiter \cite{reiter2018structured} has conducted a structural review of the widely used evaluation metric, the BLEU score. They review the evidence on how well the BLEU metric matches human evaluations. They find that BLEU is mostly valid for comparing machine translation systems, but not for other types of systems or individual texts. They also point out the limitations and biases of BLEU and human evaluations and suggest that more validation studies are needed to show how BLEU relates to real-world outcomes. They recommend that researchers do not rely on BLEU as the primary evaluation technique in their papers. 
LeClair and McMillan \cite{leclair2019recommendations} performed a study about source code summarization and their use of datasets. Code summarization lacks standardized datasets, which leads to major differences in the reported results in different papers and makes it difficult to interpret and replicate the experiments. 
They also mention that splitting the dataset by function instead of by project can cause a false boost in performance due to information leakage.
Mathur et al. \cite{mathur2020tangled} did a reevaluating study on automatic machine translation evaluation metrics and how they correlate with human judgments. They argue that the current methods for evaluating metrics are not reliable because they depend on the choice and quality of the translations. They also propose a new method for comparing different systems based on how well they agree with human judgments, and how to measure the errors of accepting or rejecting systems that are better or worse than others. 
Stapleton et al. \cite{stapleton2020human} conducted a human study on how different kinds of code summaries affect the understanding and productivity of programmers. They use human-written and machine-generated summaries for Java methods and ask participants to answer questions and write code based on them. 
They discover that human-written summaries are more helpful than machine-generated ones, but the participants do not notice any difference in quality. 
Roy et al. \cite{roy2021reassessing} did an empirical investigation on how well automatic metrics, such as BLEU, METEOR, and ROUGE, can measure the quality of code summaries generated by data-driven methods. 
They find that small differences in metric scores (less than 2 points) are not reliable indicators of better summaries and that some metrics (METEOR and chrF) are more consistent with human evaluations than others (corpus BLEU). 
Liu et al.~\cite{liu2023towards} pointed out three inappropriate settings
in existing evaluation methods of TOGA and comprehensively investigated
their impacts on evaluating and understanding the bug-finding
performance of TOGA.

Different from the above studies, in this work we revisit the performance of three recent neural oracle generation models and ChatGPT on oracle generation on both static NLG metrics and dynamic test adequacy metrics.
\section{Conclusion}
\label{sec:conclusion}
This paper conducted an empirical study of existing neural oracle generation models.
We first investigated the models' performance on different NLG-based evaluation metrics.
We assessed the generated oracles' performance in their test adequacy metrics, i.e. line coverage and mutation score.
We performed a quantitative and qualitative analysis of the generation metrics and the test adequacy to find the correlation of these models and find the gaps between what is currently being used and what we should aim to achieve from the study of this field.
We found the correlation between generation metrics and the test adequacy metrics was found not significant, meaning the currently assess generation metrics, i.e. BLEU, accuracy, Rouge-L, METEOR, and CodeBLEU, have no significant relationship with test adequacy metrics, which we use to assess the effectiveness of test cases.
This shows that NLG-based metrics are not good metrics to show the quality of the generated oracles and that test adequacy metrics should be considered the main evaluation metrics in this field.

\clearpage

\balance
\bibliographystyle{IEEEtran}
\bibliography{main}
\end{document}